\documentclass{article}

\usepackage[preprint]{neurips_2022}

\usepackage[utf8]{inputenc} 
\usepackage[T1]{fontenc}    
\usepackage{hyperref}       
\usepackage{url}            
\usepackage{booktabs}       
\usepackage{amsfonts}       
\usepackage{nicefrac}       
\usepackage{microtype}      
\usepackage{xcolor}         

\hypersetup{colorlinks,linkcolor=black,anchorcolor=blue,citecolor=black}
\usepackage{amsmath}
\usepackage{amsthm}
\newtheorem{theorem}{Theorem}[section]
\newtheorem{definition}{Definition}[section]

\newtheorem{lemma}{Lemma}

\title{Tight Lower Complexity Bounds for Strongly Convex \\ Finite-Sum Optimization}

\author{%
 Min Zhang$^1$ \qquad Yao Shu$^2$ \qquad Kun He$^1$\thanks{Corresponding author.} \vspace{0.6em}\\
$^1$School of Computer Science and Technology, \\ Huazhong University of Science and Technology, China\\
$^2$School of Computing, National University of Singapore, Singapore \vspace{0.6em}\\
\{m\_zhang, brooklet60\}@hust.edu.cn,
shuyao@comp.nus.edu.sg
}

\begin{document}

\maketitle

\begin{abstract}
Finite-sum optimization plays an important role in the area of machine learning, and hence has triggered a surge of interest in recent years. 
To address this optimization problem, various randomized incremental gradient methods have been proposed with guaranteed upper and lower complexity bounds for their convergence.
Nonetheless, these lower bounds rely on certain conditions: deterministic optimization algorithm, or fixed probability distribution for the selection of component functions. Meanwhile, some lower bounds even do not match the upper bounds of the best known methods in certain cases.
To break these limitations, we derive tight lower complexity bounds of randomized incremental gradient methods, including SAG, SAGA, SVRG, and SARAH, for two typical cases of finite-sum optimization. Specifically, our results tightly match the upper complexity of Katyusha or VRADA when each component function is strongly convex and smooth, and tightly match the upper complexity of SDCA without duality and of KatyushaX when the finite-sum function is strongly convex and the component functions are average smooth.
\end{abstract}

\section{INTRODUCTION}
Finite-sum optimization, as known as Empirical Risk Minimization, is a key problem in the area of machine learning to help the learning models achieve satisfying performance, and therefore it requires a thorough study. Specifically, we consider minimizing the following finite-sum optimization problem:
\begin{equation}
\label{obj}
\min_{x\in\mathbb{R}^d} F(x)=\frac{1}{n}\sum_{i=1}^{n}f_i(x).
\end{equation}
In this work, we mainly focus on the study of randomized incremental gradient methods with the access to Incremental First-order Oracle (IFO) for the component functions \citep{agarwal2015lower}. Formally, for $x\in\mathbb{R}^d$ and index $i\in\{1,2,\ldots,n\}$, IFO returns
\begin{equation}
h_F(x,i)=[f_i(x),\nabla f_i(x)].
\end{equation}
Note that while SAG \citep{schmidt2017minimizing}, SAGA \citep{defazio2014saga}, SVRG \citep{johnson2013accelerating,zhang2013linear} and Katyusha \citep{allen2017natasha} are included in randomized incremental gradient methods, certain dual coordinate methods such as SDCA \citep{shalev2013stochastic} are excluded and therefore areh out of the scope of this paper.

The complexity of randomized incremental gradient algorithms for strongly convex functions is literally defined as the number of IFO queries required to find an
$\epsilon$-suboptimal solution $\hat{x}$, i.e., to achieve 
\begin{equation}
F(\hat{x})-\min_{x\in\mathbb{R}^d}F(x)\leq\epsilon.
\end{equation}
For consistency, we follow this definition for the following analysis of the complexity of randomized incremental gradient algorithms.

In the literature, great efforts have been devoted to devising randomized incremental gradient algorithms under various conditions, and developing analysis of upper complexity bounds for these methods. Nevertheless, it is still important to figure out whether these methods can enjoy a smaller complexity in certain cases 
and whether there exist some other methods achieving higher performance. To answer these questions, we attempt to derive tight lower complexity bounds for the problem defined in \autoref{obj}.

There are many incremental methods and lower bounds for convex finite-sum optimization,
such as SVRG, SAGA and SAG. Algorithms based on variance-reduced stochastic gradients can be accelerated using Nesterov's momentum method \citep{nesterov2013introductory}. 
When $f_i(x)$ is convex and $L$-smooth, Katyusha, a modified Nesterov's momentum acceleration of SVRG, converges to an $\epsilon$-suboptimal solution in $O(n\log(1/\epsilon)+\sqrt{nL/\epsilon})$ times of gradient evaluation. The lower complexity bound provided by \citet{woodworth2016tight} is $\Omega(n+\sqrt{nL/\epsilon})$. When $F(x)$ is convex and $\{f_i\}_{i=1}^{n}$ is $L$-average smooth (see Definition \ref{def2}), KatyushaX, another accelerated variant of SVRG, can achieve an $\epsilon$-suboptimal solution in $O(n+n^{3/4}\sqrt{L/\epsilon})$ times of gradient computation. A lower bound is then declared by \citet{zhou2019lower} and matches the upper complexity of KatyushaX under this condition. In this work, we focus on two important cases, as described in the following.

\subsection{Two Cases to Study}

Two cases of finite-sum optimization are investigated: (1) $f_i(x)$ is $\mu$-strongly convex and $L$- smooth, i.e., $f_i(x)$ is $(\mu,L)$-smooth (see Definition \ref{def1}); (2) $F(x)$ is $\mu$-strongly convex and $\{f_i\}_{i=1}^n$ is $L$-average smooth. 


\textbf{Case (1):}  When $f_i(x)$ is $(\mu,L)$-smooth, SVRG \citep{johnson2013accelerating}, SAGA \citep{defazio2014saga}, SAG \citep{schmidt2017minimizing}, SDCA without duality \citep{shalev2016sdca} and SARAH \citep{nguyen2017sarah} can find an $\epsilon$-suboptimal solution in $O((n+L/\mu)\log(1/\epsilon))$ IFO queries, while Gradient Descent (GD) and AGD \citep{nesterov2013introductory} require $O(nL/\mu\log(1/\epsilon))$ and $O(n\sqrt{L/\mu}\log(1/\epsilon))$ IFO queries respectively. APPA \citep{frostig2015regularizing} and Catalyst \citep{lin2015universal} further reduce the IFO calls to $O((n+\sqrt{nL/\mu})\log(L/\mu)\log(1/\epsilon))$ but involve a factor $\log(L/\mu)$. By eliminating this log factor, Katyusha enjoys an upper complexity bound of $O((n+\sqrt{nL/\mu})\log(1/\epsilon))$. Regarding the lower complexity bound, \citet{agarwal2015lower} prove that achieving an $\epsilon$-suboptimal solution expects at least $\Omega(n+\sqrt{nL/\mu}\log(1/\epsilon))$ IFO calls. 
 

However, aforementioned lower bound relies on deterministic optimization algorithms and cannot be directly applied to randomized incremental gradient methods. Recently, \citet{lan2018optimal} derive a similar lower bound $\Omega(n+\sqrt{nL/\mu}\log(1/\epsilon))$ for a class of randomized algorithms where component functions are selected from predefined probability distribution. A lower bound  $\Omega(n+\sqrt{nL/\mu}\log(1/\epsilon))$ is then proposed by \citet{woodworth2016tight} and \citet{arjevani2016dimension} in a more general setting. \citet{hannah2018breaking} further improve upon the bounds in \citet{arjevani2016dimension} by bringing a lower bound $\Omega(n+\frac{n}{1+(\log(n/\kappa))_+}\log(1/\epsilon))$ for $\kappa=O(n)$ where $\kappa=L/\mu$, which matches the upper bound of VRADA (\citet{SongJM20}).


\textbf{Case (2):} When $F(x)$ is $\mu$-strongly convex, most of the existing works, including the following mentioned algorithms, assume that $f_i(x)$ is $L$-smooth. This assumption is stronger than the average smoothness assumption (see Section \ref{Pre}) and consequently can be replaced with the latter one while maintaining the upper complexity bounds as shown in \citep{zhou2019lower} . Classical SVRG method converges to an $\epsilon$-suboptimal solution with $O((n+\sqrt{n}L/\mu)\log(1/\epsilon))$ IFO calls \citep{allen2018katyusha}, while Catalyst on SVRG requires $O((n+n^{3/4}\sqrt{L/\mu})\log^2(1/\epsilon))$ IFO calls. KatyushaX \citep{allen2018katyusha} and SDCA without duality attain an upper bound of  $O((n+n^{3/4}\sqrt{L/\mu})\log(1/\epsilon))$. In terms of the lower complexity bound, \citet{zhou2019lower} reveal that any linear-span first-order randomized algorithm needs at least $\Omega(n+n^{3/4}\sqrt{L/\mu}\log(1/\epsilon))$ IFO calls.


To bridge the gap between the upper complexity bound of existing algorithms and the aforementioned lower bounds, we construct adversarial functions and provide tight lower bounds for general randomized incremental gradient methods. Therefore, our tight lower bounds suit a substantial number of optimization algorithms, e.g., SVRG, SAGA, SAG, SDCA without duality, Katyusha and KatyushaX. The upper and our improved lower complexity bounds are provided in Table \ref{bounds}.
Notably, when $n=O(L/\mu)$ in \textbf{Case (1)}, the second term $\sqrt{nL/\mu}\log(1/\epsilon)$ dominates the complexity not only in our results, but also in the lower bound derived by \citet{woodworth2016tight,arjevani2016dimension} and upper bound by \citet{katyusha}. Nonetheless, randomized incremental gradient algorithms, e.g., SVRG, SARAH and Katyusha, enjoy a  superior complexity when $n=O(L/\mu)$ and do not ignore the term $n\log(1/\epsilon)$, and our proof is more straightforward compared with previous works \citep{woodworth2016tight,zhou2019lower}. For $n \gg L/\mu$, \citet{hannah2018breaking} has proved a lower complexity bound, which matches the upper bound of their modified SVRG. We obtain this lower bound in another approach compared with them. Meanwhile, both of our analyses are in a unified framework. Similar statements also suit for \textbf{Case (2)}.

\begin{table*}[t]
\caption{Comparison between Lower Bounds and Upper Bounds on the Number of IFO Queries.}
\label{bounds}
\vspace{0.2em}
\begin{center}
\renewcommand\arraystretch{1.5}
\begin{tabular}{|c|c|c|}
\hline
 & $f_i$ is $(\mu,L)$-smooth & \begin{tabular}{c}$F(x)$ is $\mu$-strongly convex\\ and $\{f_i\}_{i=1}^n$ is $L$-average smooth\end{tabular} \\ 
\hline
 Lower bounds & \begin{tabular}{c} $n = O(\frac{L}{\mu}): \Omega((n+\sqrt{\frac{nL}{\mu}})\log\frac{1}{\epsilon})$ \\
 $n \gg \frac{L}{\mu}: \Omega(n+\frac{n}{\log(n\mu/L)}\log\frac{1}{n\epsilon})$
 \\ Theorem \ref{res1}\end{tabular} & \begin{tabular}{c}$\Omega((n+n^{3/4}\sqrt{\frac{L}{\mu}})\log\frac{1}{\epsilon})$\\ Theorem \ref{res2} \end{tabular} \\
\hline
Upper bounds & \begin{tabular}{c} $n = O(\frac{L}{\mu}): O((n+\sqrt{\frac{nL}{\mu}})\log\frac{1}{\epsilon})$ \\ \citet{katyusha} \\
$n \gg \frac{L}{\mu}: O(n+\frac{n}{\log(n\mu/L)}\log\frac{1}{n\epsilon})$
\\ \citet{SongJM20,hannah2018breaking}
\end{tabular} & \begin{tabular}{c} $O((n+n^{3/4}\sqrt{\frac{L}{mu}})\log\frac{1}{\epsilon})$ \\ \citet{shalev2016sdca}\\\citet{allen2018katyusha}\end{tabular} \\
\hline
\end{tabular}
\end{center}
\end{table*}\textbf{}

\subsection{Our Contributions}
We focus on the above two cases and our contributions are summarized as follows:
\begin{itemize}
\item When $f_i(x)$ is $(\mu,L)$-smooth and $n = O(L/\mu)$, a tight lower bound $\Omega((n+\sqrt{nL/\mu})\log(1/\epsilon))$ is derived, closely matching the upper complexity of Katyusha. Compared with the lower bound $\Omega(n+\sqrt{nL/\mu}\log(1/\epsilon))$ provided by \citet{arjevani2016dimension} and \citet{woodworth2016tight}, our lower bound reveals an optimal dependency on $n$, $L/\mu$ and $\epsilon$. In the case with $n \gg L/\mu$, we get a lower bound $\Omega(n+\frac{n}{\log(n\mu/L)}\log(1/n\epsilon))$, which matches the upper bound of \citet{SongJM20} .
\item When $F(x)$ is $\mu$-strongly convex and $\{f_i\}_{i=1}^n$ is $L$-average smooth, a tight lower bound $\Omega((n+n^{3/4}\sqrt{L/\mu})\log(1/\epsilon))$ is obtained for any randomized incremental gradient method, tightly matching the upper bound of SDCA without duality and of KatyushaX. Compared with the lower bound $\Omega(n+n^{3/4}\sqrt{L/\mu}\log(1/\epsilon))$ derived by \citet{zhou2019lower}, our results also affirm an optimal dependency on $n$, $L/\mu$ and $\epsilon$.
\item We expose that the quality of the solution $\epsilon$ is independent on $n$ and $L/\mu$, while \citet{woodworth2016tight} require $\epsilon=O(\sqrt{nL/\mu})$ for Case (1) and \citet{zhou2019lower} demand $\epsilon=O(n^{7/4}(L/\mu)^{-3/2})$ for Case (2).
\item Concerning the analytical techniques, the adversarial functions we construct are straightforward and our analysis is established on the distance between $\hat{x}$ and the optimal $x^*$, i.e. $\|\hat{x}-x^*\|$. We, therefore, are capable to achieve a lower complexity bound of $\hat{x}$ such that $\|\hat{x}-x^*\|\leq\epsilon$. Furthermore, without the assumption of predefined distribution for the selection of component functions and the limitation of the calls of component functions, our lower bounds are derived under a more general setting, as revealed in the proof.
\end{itemize}

\section{PRELIMINARIES}
\label{Pre}

We begin our analysis with the definitions of convexity and smoothness of functions, and Nesterov's chain-like quadratic function,  which is the foundation of adversarial functions. 

We first recall the definitions of convexity, strong convexity and smoothness.
\begin{definition}
\label{def1}
For a differentiable function $f$: $\mathbb{R}^d \rightarrow \mathbb{R}$, 
\begin{itemize}
\item $f$ is convex if $\forall x, y \in \mathbb{R}^d$,  it satisfies 
\[ f(y) \geq f(x) +\langle \nabla f(x), y-x \rangle. \]
\item $f$ is $\mu$-strongly convex if for some $\mu>0$, $\forall x,  y \in \mathbb{R}^d$, it satisfies 
\[ f(y) \geq f(x) +\langle \nabla f(x), y-x \rangle + \frac{\mu}{2}\|y-x\|_2^2.\]
\item $f$ is $L$-smooth if for some $L>0$, $\forall x, y \in \mathbb{R}^d$, it satisfies
\[\|\nabla f(x)-\nabla f(y)\|_2 \leq L\|x-y\|_2.\]

\item $f$ is $(\mu, L)$-smooth if for some $\mu, L>0$, $\forall x, y \in \mathbb{R}^d$, it satisfies
\[
\begin{aligned}
\frac{\mu}{2}\|y-x\|_2^2 &\leq f(y)-f(x)-\langle \nabla f(x), y-x \rangle
\leq \frac{L}{2}\|y-x\|_2^2.
\end{aligned}
\]
\end{itemize} 
\end{definition} 

Note that for a twice differentiable function $f$, it is convex if and only if all eigenvalues of $\nabla^2f(x)$ are non-negative; $f$ is $\mu$-strongly convex if and only if all eigenvalues are at least $\mu$; $f$ is $L$-smooth if and only if all eigenvalues are no more than $L$; $f$ is $(\mu, L)$-smooth if and only if all eigenvalues are no less than $\mu$ and no more than $L$. 

\begin{definition}
\label{def2}
 For any sequence of differentiable functions $\{f_i\}_{i=1}^n$:$f_i: \mathbb{R}^d\rightarrow\mathbb{R} $,  we say $\{f_i\}_{i=1}^n$ is $L$-average smooth if $\forall x, y \in \mathbb{R}^d$, it satisfies
 \begin{equation}
\frac{1}{n}\sum_{i=1}^{n}\|\nabla f_i(x)-\nabla f_i(y)\|_2^2 \leq L^2\|x-y\|_2^2.
\end{equation}
\end{definition}
The average smoothness assumption appears in previous works \citep{zhou2018stochastic,fang2018spider,zhou2019lower,xie2019general}, which is adopted by \citet{zhou2019lower} and \cite{xie2019general} to prove lower bounds for finite-sum optimization. It can be easily verified that if each $f_i$ is $L$-smooth,  then $\{f_i\}_{i=1}^n$ is $L$-average smooth. If $\{f_i\}_{i=1}^n$ is $L$-average smooth,  with the convexity of $\|\bullet\|_2^2$ 
we can derive that $F(x)$ defined in \autoref{obj} is $L$-smooth \footnote{The convexity of $\|\bullet\|_2^2$ implies that 
$\|\nabla F(x)-\nabla F(y)\|_2^2=\|\frac{1}{n}\sum_{i=1}^{n}\nabla f_i(x)-\frac{1}{n}\sum_{i=1}^{n}\nabla f_i(y)\|_2^2=\|\frac{1}{n}\sum_{i=1}^{n} \left(\nabla f_i(x)-\nabla f_i(y) \right)\|_2^2 \leq \frac{1}{n}\sum_{i=1}^{n}\|\nabla f_i(x)-\nabla f_i(y)\|_2^2 \leq L^2\|x-y\|_2^2$, and so we have $\|\nabla F(x)-\nabla F(y)\|_2 \leq L\|x-y\|_2$.}.

In order to construct adversarial functions to help introduce our lower bounds, we first introduce the following classes of quadratic functions \citep{nesterov2013introductory}, which were used to prove lower bounds for smooth strongly convex optimization problems. 

Let us choose $\mu>0$ and $L\geq\mu$. The function defined in the infinite-dimensional space $\mathbb{R}^\infty$ is given by
\begin{equation}
f(x)=\frac{L-\mu}{4}\left(\frac{1}{2}\langle x,Ax\rangle-\langle e_1,x\rangle \right)+\frac{1}{2}\mu\|x\|_2^2, 
\end{equation}
where 
\begin{equation}
A:=
\begin{pmatrix}
2 & -1 & 0 & \\
-1 & 2 & -1 & 0 \\
0 & -1 & 2 & \\
 & 0 &  &\cdots
\end{pmatrix}
\end{equation}
and $e_1$ is the unit vector in which the first entry is $1$. Note that for any $s=(s_1,s_2, \ldots)^T\in \mathbb{R}^\infty$, we have
\[
    \langle s,As\rangle = s_1^2+\sum_{i=1}^{\infty}(s_i-s_{i+1})^2 \geq 0,
\]
and
\[
\begin{aligned}
    \langle s,As\rangle &\leq s_1^2+\sum_{i=1}^{\infty}2(s_i^2+s_{i+1}^2) 3s_1^2+\sum_{i=2}^{\infty}4s_i^2 \leq 4\sum_{i=1}^{\infty}s_i^2.
\end{aligned}
\]
Thus, $0 \preceq A \preceq 4I_\infty$, where $I_\infty$ is the unit matrix in $\mathbb{R}^\infty$. We can see that $\nabla^2f(x)=\frac{L-\mu}{4}A+\mu I_\infty$. Therefore, $f(x)$ is $L$-smooth and $\mu$-strongly convex. Applying first order optimality condition $\nabla f(x)=0$, we can get $x^*=(q,q^2,q^3,\ldots)$, where $q=\frac{\sqrt L-\sqrt\mu}{\sqrt L+\sqrt\mu}$ \citep{nesterov2013introductory}.

In fact, function $f(x)$ is a first-order zero-chain \citep{carmon2019lower}, which implies for any vector $x=(x_1,x_2,\ldots)\in\mathbb{R}^\infty$ where only the first $k-1$ entries are nonzero, i.e., $x_{k}=x_{k+1}\ldots=0$, we have $\nabla_i f(x)=0$ for any $i\geq k+1$. In other words, the information brought by a query to IFO can increase at most one nonzero element. We leverage this chain structure to prove lower bounds in our framework. Meanwhile, note that this function can also be defined in finite-dimensional space with some modifications, which is reflected in the construction of adversarial function in Section \ref{sec:Proof}.

\section{MAIN RESULTS}
\label{sec_results}
Here we present our lower bounds of randomized incremental gradient algorithms. We first establish lower bound for the sum of $(\mu,L)$-smooth functions, then turn to functions where $\{f_i\}_{i=1}^n$ is $L$-average smooth and $F(x)$ is $\mu$-strongly convex.

\begin{theorem}
\label{res1}
For any $L, \mu >0$, $n\geq2$, $0<\epsilon<\frac{B^2\mu}{4}$, and any randomized incremental gradient method $\mathcal{M}$, there exist a dimension $d=O((n+\sqrt{n\frac{L}{\mu}}) \log \frac{1}{\epsilon})$ in the case with $\frac{L}{\mu} > \frac{7}{2}n$ or $d=O(n+\frac{n}{\log\frac{n\mu}{L}} \log \frac{1}{n\epsilon})$ in the case with $n \gg L/\mu$, $x^0\in \mathbb{R}^d$ and n $(\mu,L)$-smooth functions $\{f_i\}_{i=1}^n$: $\mathbb{R}^d\rightarrow \mathbb{R}$, such that $\|x^0-x^*\|=B$ where $x^*=\mathop{\arg\min}_{x \in\mathbb{R}^d}F(x)$. In order to find $F(\hat{x})-F(x^*)\leq\epsilon$, $\mathcal{M}$ needs at least
\begin{equation}
\Omega((n+\sqrt{\frac{nL}{\mu}}) \log \frac{B^2\mu}{4\epsilon})
\end{equation}
IFO queries when $\frac{L}{\mu} > \frac{7}{2}n$, and needs at least
\begin{equation}
\Omega(n+\frac{n}{\log\frac{n\mu}{L}} \log \frac{1}{n\epsilon})
\end{equation}
IFO queries when $n \gg \frac{L}{\mu}$
\end{theorem}
The lower bound in Theorem \ref{res1} tightly matches the upper bound of Katyusha \citep{katyusha} when each component function is $(\mu,L)$-smooth. Compared with the complexity bound for strongly convex and smooth finite-sum optimization derived by \citet{woodworth2016tight}, who prove $\Omega(n+\sqrt{n\kappa}\log\frac{1}{\epsilon})$ lower bound, our results provide optimal dependency on $n$, $\kappa$ and $\epsilon$. As a side result, the accuracy of solution $\epsilon$ is independent on $n$ compared with their work. Under the second condition, our lower bound also matches the upper bound of VRADA (\citet{SongJM20}) and modified SVRG (\citet{hannah2018breaking})

Then we give the lower bound when $F(x)$ is strongly convex and $\{f_i\}_{i=1}^n$ is average smooth.
\begin{theorem}
\label{res2}
For any $L, \mu >0$, $n\geq2$ such that $\frac{L}{\mu}\geq \frac{9}{4}\sqrt{n}$, any $0<\epsilon<\frac{B^2\mu}{4}$, and any randomized incremental gradient method $\mathcal{M}$, there exist a dimension $d=O((n+n^{\frac{3}{4}}\sqrt{\frac{L}{\mu}}) \log \frac{1}{\epsilon})$, $x^0\in\mathbb{R}^d$ and functions $\{f_i\}_{i=1}^n$: $\mathbb{R}^d\rightarrow \mathbb{R}$ where $\{f_i\}_{i=1}$ is $L$-average smooth and $F(x)$ is $\mu$-strongly convex, such that $\|x^0-x^*\|=B$ where $x^*=\mathop{\arg\min}_{x \in\mathbb{R}^d}F(x)$. In order to find $F(\hat{x})-F(x^*)\leq\epsilon$, $\mathcal{M}$ needs at least
\begin{equation}
\Omega ((n+n^\frac{3}{4}\sqrt{\frac{L}{\mu}}) \log \frac{B^2\mu}{4\epsilon})
\end{equation}
IFO queries.
\end{theorem}
The lower bound in Theorem \ref{res2} under the first condition also tightly matches the upper bound of SDCA without duality \citep{shalev2016sdca} and of KatyushaX \citep{allen2018katyusha} when $F(x)$ is $\mu$-strongly convex. While \cite{zhou2019lower} provide $\Omega(n+n^\frac{3}{4}\sqrt{\kappa}\log\frac{1}{\epsilon})$ lower bound and our result provides optimal dependency on $n$, $\kappa$ and $\epsilon$. Furthermore, in our result, the accuracy of solution $\epsilon$ has no dependency on $n$.

\textbf{Remark} The lower complexity bound in Theorem \ref{res2} relies on the average smoothness assumption. 
In fact, many existing finite-sum optimization algorithms such as SDCA without duality \citep{shalev2016sdca}, Natasha \citep{allen2017natasha}, KatyushaX \citep{allen2018katyusha}, RapGrad \citep{lan2019accelerated} and StagewiseKatyusha \citep{chen2018variance} assume that $f_i$ is $L$-smooth and we have claimed that this assumption is stronger than the average smoothness. Previous work \citep{zhou2019lower} indicates that we can replace the smoothness assumption of these algorithms with average smoothness without affecting their upper complexity bounds, and therefore, our average smoothness assumption is reasonable.

\section{PROOFS OF MAIN RESULTS}
\label{sec:Proof}
In this section, the proofs of our main results are provided and the proof of lemmas are in the Appendix.
\subsection{Proof of Theorem \ref{res1}}
\label{subsec:proof_thm1}
We construct a special class of finite-sum optimization problem with the form of
\begin{align}
\label{obj1}
\min_{x \in \mathbb{R}^d} F(x)
=&\frac{1}{n}\sum_{i=1}^{n}f_i(x) \nonumber \\ 
=&\frac{1}{n}\sum_{i=1}^{n}\left[\frac{L-\mu}{4}\left(\frac{1}{2}\langle x,A_i x\rangle-
  \langle e_{(i-1)p+1},x\rangle\right)\right. +\left.\frac{1}{2}\mu\|x\|_2^2\right],
\end{align}
where $p=\frac{d}{n}\in\{1,2,3\ldots\}$ and $e_{k}$ is a unit vector in which the $k$-th element is 1. Let $A_i$ be a symmetric matrix in $\mathbb{R}^{d \times d}$ defined as
\begin{equation*}
\label{matrix}
\begin{pmatrix}
\begin{smallmatrix}
 0_{p(i-1),p(i-1)} & 0_{p(i-1),p} & 0_{p(i-1),d-pi} \\[1.5em]
 & \left.\begin{matrix}
 \begin{smallmatrix}
 \begin{smallmatrix}2 & -1 & 0 &\\ -1 & 2 & -1 \\0 &-1& 2 \end{smallmatrix} & \\
 \cdots & \cdots \\
  & \begin{smallmatrix} -1 & 2 & -1\\ 0 & -1 & \xi \end{smallmatrix} 
\end{smallmatrix}
 \end{matrix}\right\}p \times p & \\[1.5em]
 0_{d-pi,p(i-1)} &  0_{d-pi,p} &  0_{d-pi,d-pi}
 \end{smallmatrix}
\end{pmatrix}
, 
\end{equation*}
 where $0_{k,p}$ is $(k\times p)$ zero matrix and $\xi=\frac{\left(\frac{\kappa-1}{n}+1\right)^{\frac{1}{2}}+3}{\left(\frac{\kappa-1}{n}+1\right)^{\frac{1}{2}}+1}$ in which $\kappa=\frac{L}{\mu}$. 

It's easy to see that $\nabla^2f_i(x)=\frac{L-\mu}{4}A_i + \mu I$, where $I$ is the unit matrix in $\mathbb{R}^{d \times d}$. Due to the fact that $0 \preceq A_i \preceq 4I $, $\mu I \preceq \nabla^2f_i(x) \preceq LI$.  Thus, $f_i(x)$ is $(\mu,L)$-smooth. The following lemma gives an explicit expression of the minimizer in \autoref{obj1}.
\begin{lemma}
\label{lemma:minimizer_case1}
Let $x^*$ be the minimizer of function $F(x)$ in \autoref{obj1}, and
\begin{equation}
q_1=q_2\ldots=q_n=\frac{\left(\frac{\kappa-1}{n}+1\right)^{\frac{1}{2}}-1}{\left(\frac{\kappa-1}{n}+1\right)^{\frac{1}{2}}+1},
\end{equation}
where $\kappa=\frac{L}{\mu}$, then $x^*=(q_1,q_1^2, \ldots,q_1^p,q_2,q_2^2,\ldots,q_2^p,
\ldots,q_n,q_n^2,\ldots, q_n^p)^T$.
\end{lemma}

Without loss of generality, we assume that the initial point $x^0=0$, otherwise we can take $\hat{F}(x)=F(x-x^0)$. The property of function $f_i$ implies that every query to the gradient of component function $f_i(x)$ can only ``discover'' a new coordinate.
Let $x^K$ be the point generated after $K$ gradient evaluations (e.g., the point generated by one full gradient iteration of $F(x)$ is denoted as $x^n$), and $K_i \in\{0,1,\ldots,K\}$ be the query times of $\nabla f_i$, with $K=K_1+K_2+\ldots K_n$. Note that $K_i$ queries to $\nabla f_i$ result in at most $K_i$ coordinates of nonzero elements in $x^K$. For simplicity, we divide $x^K$ into $n$ parts and let $x^K=(x_{1}^{K},x_{2}^{K},\ldots, x_{n}^{K})$, in which $x_i^K$ ($i \leq n$) has $p$ elements. 

Let $S=\{i \in \{1,2,\ldots,n\}|K_i<p\}$ and $S^c=\{i \in \{1,2,\ldots,n\}|K_i\geq p\}$. It means that our proof also considers the case where the algorithms query some component functions more than $p$ times. Assuming $K \leq \frac{d}{2}=\frac{np}{2}$, and then we can establish the lower bound on how close $x^K$ is to $x^*$:
\begin{align}
\label{xkx*}
 \frac{\|x^K-x^*\|_2^2}{\|x^0-x^*\|_2^2} &=\frac{\sum_{i=1}^{n}\|x_i^K-(q_i,q_i^2,\ldots,q_i^p)\|_2^2}{n \|(q_1,q_1^2,\ldots,q_1^p)\|_2^2} \nonumber \\
 &\geq \frac{\sum_{i\in S}\sum_{j=K_i+1}^{p} q_1^{2j}}{\frac{nq_1^2(1-q_1^{2p})}{1-q_1^2}} \nonumber \\
 &=\frac{\sum_{i\in S}(q_1^{2K_i}-q_1^{2p})}{n(1-q_1^{2p})} \nonumber \\
 &\geq \frac{\sum_{i\in S}(q_1^{2K_i}-q_1^{2p})+\sum_{i\in S^c}(q_1^{2K_i}-q_1^{2p})}{n(1-q_1^{2p})} \nonumber \\
 &=\frac{\sum_{i=1}^{n}(q_1^{2K_i}-q_1^{2p})}{n(1-q_1^{2p})}.
\end{align}
The first inequality in \autoref{xkx*} follows the property that $x_i^K$ has at most $\min(K_i,p)$ nonzero elements. The second inequality using the following fact: for any $i\in S^c$, $q_1^{2K_i}\leq q_1^{2p}$. Note that $q_1^x$ is convex with respect to $x$, and then we have
\begin{align}
\frac{\|x^K-x^*\|_2^2}{\|x^0-x^*\|_2^2} &\geq \frac{\frac{1}{n}\sum_{i=1}^{n}q_1^{2K_i}-q_1^{2p}}{1-q_1^{2p}} \nonumber\\
&\geq \frac{q_1^{\frac{1}{n}\sum_{i=1}^{n}2K_i}-q_1^{2p}}{1-q_1^{2p}} \nonumber\\
&= \frac{q_1^{\frac{2K}{n}}-q_1^{2p}}{1-q_1^{2p}}. 
\end{align}
Observing that $\frac{2K}{n}\leq p$, it implies that
\begin{align}
\frac{\|x^K-x^*\|_2^2}{\|x^0-x^*\|_2^2} &\geq \frac{q_1^{\frac{2K}{n}}-q_1^{2p}}{1-q_1^{2p}}
\geq \frac{q_1^{\frac{2K}{n}}-q_1^{p+\frac{2K}{n}}}{1-q_1^{2p}} \nonumber\\
&=\frac{q_1^{\frac{2K}{n}}(1-q_1^p)}{1-q_1^{2p}} \nonumber\\
&=\frac{q_1^{\frac{2K}{n}}}{1+q_1^p}.
\end{align}
Since $q_1<1$, we have
\begin{align}
\|x^K-x^*\|_2^2 &\geq\frac{1}{1+q_1^p}\|x^0-x^*\|_2^2 \geq \frac{1}{2}q_1^{\frac{2K}{n}}\|x^0-x^*\|_2^2 \frac{B^2}{2}q_1^{\frac{2K}{n}}.
\end{align}
Above inequality implies that, in addition to $F(x^K)-F(x^*)\leq\epsilon$, our proof can also obtain a lower bound when $\|x^K-x^*\|\leq\epsilon\|x^0-x^*\|$ \footnote{Letting $\|x^K-x^*\|\leq\epsilon\|x^0-x^*\|$, we can also establish a lower bound with respect to the number of queries $K$ and it could be a side result. }. Noting that $F(x)$ is $\mu$-strongly convex, by the property of strongly convexity, it is easy to get
\begin{align}
F(x^K)-F(x^*) &\geq \frac{\mu}{2}\|x^K-x^*\|_2^2 \geq \frac{B^2\mu}{4}q_1^{\frac{2K}{n}}.
\end{align}
In order to get $F(x^K)-F(x^*) \leq \epsilon$, $K$ must satisfy 
\begin{equation}
K \geq n\frac{\log \frac{B^2\mu}{4\epsilon}}{\log \frac{1}{q_1}}.
\end{equation}

We now provide an estimation on $\frac{1}{\log\frac{1}{q_1}}$ for $n=O(\kappa)$.
\begin{lemma}
\label{lemma:log_case1}
Given  $\kappa>1$, $n\geq2$ and $q_1=\frac{\left(\frac{\kappa-1}{n}+1\right)^{\frac{1}{2}}-1}{\left(\frac{\kappa-1}{n}+1\right)^{\frac{1}{2}}+1}$, if $\kappa \geq \frac{7}{2}n$, we have
\begin{equation}
\frac{1}{\log\frac{1}{q_1}} \geq \frac{1}{2\sqrt{14}}\sqrt{\frac{\kappa}{n}}+\frac{1}{4}.
\end{equation}
\end{lemma}

In the case with $n \gg \kappa$, we provide the following lemma.
\begin{lemma}
\label{lemma:log_case11}
Given $B, L,\mu>0, \kappa=L/\mu>1$, $n\geq2$ and $q_1=\frac{\left(\frac{\kappa-1}{n}+1\right)^{\frac{1}{2}}-1}{\left(\frac{\kappa-1}{n}+1\right)^{\frac{1}{2}}+1}$, if $n \gg \kappa$, we have
\begin{equation}
\frac{\log \frac{B^2\mu}{4\epsilon}}{\log \frac{1}{q_1}} \geq c + \frac{1}{\log\frac{n\mu}{L}}\log\frac{1}{n\epsilon},
\end{equation}
where $c$ is some constant.
\end{lemma}

Thus, it can be derived that for any $\epsilon <\frac{B^2\mu}{4}$ and $\kappa \geq \frac{7}{2}n$, the number of gradient evaluations needed is at least
\begin{equation}
K = \Omega ((n+\sqrt{n\kappa}) \log \frac{B^2\mu}{4\epsilon}).
\end{equation}
Noting that $K\leq\frac{d}{2}$, we can see that there must exist a dimension $d=O((n+\sqrt{n\kappa}) \log \frac{1}{\epsilon})$ that satisfies the above statement~\footnote{We hide unimportant factors in logarithmic term.}.

In the case with $n \gg \kappa$, we can obtain
\begin{equation}
K = \Omega(n+\frac{n}{\log\frac{n\mu}{L}} \log \frac{1}{n\epsilon}),
\end{equation}
and $d=O(n+\frac{n}{\log\frac{n\mu}{L}} \log \frac{1}{n\epsilon})$.

\subsection{Proof of Theorem \ref{res2}}
\label{subsec:proof_thm2}
We also construct a special class of finite-sum optimization problem with the form of
\begin{align}
\label{obj2}
\min_{x \in \mathbb{R}^d} F(x) = &\frac{1}{n}\sum_{i=1}^{n}f_i(x) \nonumber\\
=&\frac{1}{n}\sum_{i=1}^{n}\left[\frac{\sqrt{n}L-n\mu}{4}\right.
\left.\left(\frac{1}{2}\langle x,A_i x\rangle-\langle e_{(i-1)p+1},x\rangle\right) + \frac{n\mu}{2}\langle x,B_ix\rangle\right],
\end{align}
where $A_i$ has the same form as in Section \ref{subsec:proof_thm1} with $\xi=\frac{\sqrt{\kappa}+3n^{\frac{1}{4}}}{\sqrt{\kappa}+n^{\frac{1}{4}}}$, in which $\kappa=\frac{L}{\mu}>\sqrt{n}$ to guarantee $\sqrt{n}L-n\mu >0 $, and $B_i$ is a symmetric matrix in $\mathbb{R}^{d \times d}$ defined as
\begin{equation*}
\begin{pmatrix}
0_{p(i-1),p(i-1)} & \\[1em]
&I_{p,p} & \\[1em]
&  & 0_{d-pi,d-pi}
\end{pmatrix},
\end{equation*}
where $0_{m, n}$ is an $(m\times n)$ zero matrix and $I_{p,p}$ is the unit matrix in $\mathbb{R}^{p\times p}$. Likewise, $p=\frac{d}{n}\in\{1,2,3\ldots\}$ and $e_{k}$ is a unit vector in which the $k$-th element is 1.

The lemma below describes the average smoothness of $\{f_i\}_{i=1}^n$.
\begin{lemma}
\label{lemma:smoothness_case2}
For \autoref{obj2}, $\{f_i\}_{i=1}^n$ is $L$-average smooth.
\end{lemma}
Observing that
\begin{equation*}
\begin{aligned}
\nabla^2 F(x) &=\frac{\sqrt{n}L-n\mu}{4n}\sum_{i=1}^{n}A_i+\mu\sum_{i=1}^{n}B_i \\
              &=\frac{\sqrt{n}L-n\mu}{4n}\sum_{i=1}^{n}A_i+\mu I,
\end{aligned}
\end{equation*}
where $I$ is the unit matrix in $\mathbb{R}^{d \times d} $, $F(x)$ is $\mu$-strongly convex. 
The following lemma gives the optimal solution in \autoref{obj2}.
\begin{lemma}
\label{lemma:minimizer_case2}
Let $x^*$ be the minimizer of function $F(x)$ in \autoref{obj2}, and
\begin{equation}
q_1=q_2\ldots=q_n=\frac{\sqrt{\kappa}-n^{\frac{1}{4}}}{\sqrt{\kappa}+n^{\frac{1}{4}}},
\end{equation}
where $\kappa=\frac{L}{\mu}$, then $x^*=(q_1, q_1^2, \ldots, q_1^p, q_2, q_2^2, \ldots, q_2^p, \ldots, q_n, q_n^2, \ldots, q_n^p)^T$.
\end{lemma}


Let $x^K$ be the point generated after $K$ gradient evaluations and $K_i \in\{0,1,\ldots,K\}$ be the query times of $\nabla f_i$. Using the same notations and methods, and assuming $K \leq \frac{d}{2}$, we establish the lower bound of $\|x^K-x^*\|_2^2$:
\begin{equation}
\|x^K-x^*\|_2^2 \geq \frac{B^2}{2}q_1^{\frac{2K}{n}}.
\end{equation}

By $\mu$-strongly convexity of $F(x)$, we have
\begin{align}
F(x^K)-F(x^*) &\geq \frac{\mu}{2}\|x^K-x^*\|_2^2
\geq \frac{B^2\mu}{4}q_1^{\frac{2K}{n}}.
\end{align}
In order to get $F(x^K)-F(x^*) \leq \epsilon$, $K$ must satisfy
\begin{equation}
K \geq n\frac{\log \frac{B^2\mu}{4\epsilon}}{\log \frac{1}{q_1}}.
\end{equation}

The following lemma gives an estimation on $\frac{1}{\log\frac{1}{q_1}}$.
\begin{lemma}
\label{lemma:log_case2}
Given $q_1=\frac{\sqrt{\kappa}-n^{\frac{1}{4}}}{\sqrt{\kappa}+n^{\frac{1}{4}}}$, if $\kappa \geq \frac{9}{4}\sqrt{n}$, then we have
\begin{equation}
\frac{1}{\log\frac{1}{q_1}} \geq \frac{1}{12}n^{-\frac{1}{4}}\sqrt{\kappa}+\frac{1}{8}.
\end{equation}
\end{lemma}

As a result, for any $\epsilon <\frac{B^2\mu}{4}$ and $\kappa \geq \frac{9}{4}\sqrt{n}$, the number of the gradient evaluations needed is at least
\begin{equation}
K = \Omega ((n+n^\frac{3}{4}\sqrt{\kappa}) \log \frac{B^2\mu}{4\epsilon}). 
\end{equation}
Meanwhile, we can also derive that there must exist a function $F(x)$ with dimension $d= O((n+n^\frac{3}{4}\sqrt{\kappa}) \log \frac{1}{\epsilon})$.

\section{CONCLUSION}
We have established tight lower bounds on the Incremental First-order Oracle complexity for randomized incremental gradient method in solving two important cases of finite-sum optimization problems. To some extent, our results are general for first-order primal methods. When each component function is strongly convex and smooth, a tight lower bound is obtained. For a general setting that the finite-sum function is strongly convex and the component functions are average smooth, our lower bound also tightly matches the upper complexity of existing algorithms. We should point out that in this case, the condition $L/\mu= \Omega(\sqrt{n})$ must be satisfied in our analysis framework. Thus, how to obtain the lower bound when $L/\mu = O(\sqrt{n})$ could be a future study.

\bibliography{ref}

\newpage
\appendix
\part*{Appendix}

\section{Proof of Lemmas in Section \ref{subsec:proof_thm1}}
\subsection{Proof of Lemma \ref{lemma:minimizer_case1}}
Using first order optimality condition, we obtain
\begin{equation*}
\left(\frac{L-\mu}{4}\sum_{i=1}^{n}A_i+n\mu I \right)x
-\frac{L-\mu}{4}\sum_{i=1}^{n}e_{(i-1)p+1}=0.
\end{equation*}
Rewrite this equation as
\begin{equation*}
\left(\sum_{i=1}^{n}A_i+\frac{4nI}{\kappa-1}\right)x=\sum_{i=1}^{n}e_{(i-1)p+1}.
\end{equation*}
Let $x=(x_{1,1},x_{1,2},\ldots,x_{1,p},x_{2,1}, x_{2,2},\ldots,x_{2,p},\ldots, \\ x_{n,1},x_{n,2}\ldots,x_{n,p})^T$, then for any $i\in\{1,2, \ldots,n\}$, we have
\begin{equation*}
  \begin{aligned}
\frac{2(\kappa-1+2n)}{\kappa-1}x_{i,1}-x_{i,2} &=1 ,\\
x_{i,k+1}-\frac{2(\kappa-1+2n)}{\kappa-1}x_{i,k}+x_{i,k-1} &=0, \\
\left(\xi+\frac{4n}{\kappa-1}\right) x_{i,p}-x_{i,p-1} &=0,
  \end{aligned}
\end{equation*}
where $k=2,3\ldots,p-1$. 
Note that $q_i=\frac{\sqrt{\frac{\kappa-1}{n}+1}-1}{\sqrt{\frac{\kappa-1}{n}+1}+1}$ is the smallest root of equation
\begin{equation*}
q_i^2-\frac{2(\kappa-1+2n)}{\kappa-1}q_i+1=0
\end{equation*}
and the root of equation
\begin{equation*}
  \left(\xi+\frac{4n}{\kappa-1}\right)q_i-1=0.
\end{equation*}
Then $x^*=(q_1,q_1^2, \ldots,q_1^p,q_2,q_2^2,\ldots,q_2^p,\ldots,q_n,q_n^2,\ldots,  q_n^p)^T$ satisfies the first order optimal condition, completing the proof. 

\subsection{Proof of Lemma \ref{lemma:log_case1}}
Due to the fact that $\log(1+x)\leq x$, we have
\begin{equation*}
\label{app_eq:q_1}
  \begin{aligned}
\frac{1}{\log\frac{1}{q_1}} &=\frac{1}{\log\frac{\sqrt{\frac{\kappa-1}{n}+1}+1}{\sqrt{\frac{\kappa-1}{n}+1}-1}}
=\frac{1}{\log(1+\frac{2}{\sqrt{\frac{\kappa-1}{n}+1}-1})}    \\
&\geq\frac{\sqrt{\frac{\kappa-1}{n}+1}-1}{2}  \\
&\geq \frac{\sqrt{\frac{\kappa}{n}+\frac{1}{2}}-1}{2},
  \end{aligned}
\end{equation*}
where the last inequality follows from $n\geq 2$. Since $\sqrt{x+\frac{1}{2}}-1\geq \frac{1}{\sqrt{14}}\sqrt{x}+\frac{1}{2}$ when $x\geq\frac{7}{2}$, i.e. $\frac{n}{\kappa}\geq\frac{7}{2}$, it could be derived that
\begin{equation*}
\begin{aligned}
\frac{1}{\log\frac{1}{q_1}} &\geq \frac{\sqrt{\frac{\kappa}{n}+\frac{1}{2}}-1}{2} \\
&\geq \frac{1}{2\sqrt{14}}\sqrt{\frac{\kappa}{n}}+\frac{1}{4}.
\end{aligned}
\end{equation*}
Thus, the lemma is proved.

\subsection{Proof of Lemma \ref{lemma:log_case11}}
We first estimate $\frac{1}{\log 1/q_1}$. Since $n \gg \kappa$, we have
\begin{equation*}
\label{app_eq:q_2}
  \begin{aligned}
\frac{1}{\log\frac{1}{q_1}} 
&=\frac{1}{\log\frac{\sqrt{\frac{\kappa-1}{n}+1}+1}{\sqrt{\frac{\kappa-1}{n}+1}-1}} = \frac{1}{\log\frac{(\sqrt{\frac{\kappa-1}{n}+1}+1)^2}{\frac{\kappa-1}{n}}}  \\
&= \frac{1}{\log\frac{2 + \frac{\kappa-1}{n}+  2\sqrt{\frac{\kappa-1}{n}+1}}{\frac{\kappa-1}{n}}} \\
&\ge \frac{1}{\log\frac{2+2\sqrt{2}+\frac{\kappa-1}{n}}{\frac{\kappa-1}{n}}} \\
&= \frac{1}{\log(1+\frac{(2+2\sqrt{2})n}{\kappa-1})} \\
&\ge \frac{1}{\gamma\log\frac{n}{\kappa}},
  \end{aligned}
\end{equation*}
where $\gamma$ is some constant. Then we obtain
\begin{equation*}
\label{app_eq:q_3}
  \begin{aligned}
\frac{\log\frac{B^2\mu}{4\epsilon}}{\log\frac{1}{q_1}} 
&\ge \frac{\log\frac{n}{\kappa}+\log\frac{B^2L}{n\epsilon}}{\gamma\log\frac{n}{\kappa}} \\
&= \frac{1}{\gamma} + \frac{\log\frac{B^2L}{n\epsilon}}{\gamma\log\frac{n}{\kappa}}.
  \end{aligned}
\end{equation*}
This finishes the proof.

\section{Proof of Lemmas in Section \ref{subsec:proof_thm2}}
\subsection{Proof of Lemma \ref{lemma:smoothness_case2}}
For simplicity, we set $\alpha=\frac{\sqrt{n}L-n\mu}{4}$ and $\beta=n\mu$ such that $\nabla f_i(x)=\alpha(A_ix-e_{(i-1)p+1})+\beta B_ix$. Therefore, for any $i\in\{1,2,\ldots,n\}$, we have
\begin{equation*}
\begin{aligned}
&\|\nabla f_i(x)-\nabla f_i(y)\|_2^2 \nonumber\\
=&\|\alpha A_ix+\beta B_ix-\alpha A_iy-\beta B_iy\|_2^2 \\
=&\|(\alpha A_i+\beta B_i)(x-y)\|_2^2 \\
=& (x-y)^T(\alpha^2A_i^2+2\alpha\beta A_i B_i+\beta^2B_i^2)(x-y) \\
=&(x-y)^T(\alpha^2A_i^2+2\alpha\beta A_i+\beta^2B_i)(x-y),
\end{aligned}
\end{equation*}
where 
the last equality uses two simple relations that $A_iB_i=A_i$ and $B_i^2=B_i$.

Thus, we have
\begin{equation*}
  \begin{aligned}
&\frac{1}{n}\sum_{i=1}^{n}\|\nabla f_i(x)-\nabla f_i(y)\|_2^2 \\
=&\frac{1}{n}\sum_{i=1}^{n}(x-y)^T(\alpha^2A_i^2+2\alpha\beta A_i+\beta^2B_i)(x-y) \\
=&\frac{1}{n}(x-y)^T(\alpha^2\sum_{i=1}^{n}A_i^2+2\alpha\beta A+\beta^2I)(x-y),
  \end{aligned}
\end{equation*}
where $A=\sum_{i=1}^{n}A_i$. Observing that
\begin{equation*}
A^2=(\sum_{i=1}^{n}A_i)^2=\sum_{i=1}^{n}A_i^2,
\end{equation*}
where $A_i$ is a block diagonal matrix in which only the $i$-th block is nonzero and $A_i\preceq 4I$, we can obtain $A=\sum_{i=1}^{n}A_i\preceq 4I$ and $A^2\preceq16I$.

Then we conclude that
\begin{equation*}
  \begin{aligned}
&\frac{1}{n}\sum_{i=1}^{n}\|\nabla f_i(x)-\nabla f_i(y)\|_2^2 \\
=&\frac{1}{n}(x-y)^T(\alpha^2A^2+2\alpha\beta A+\beta^2I)(x-y) \\
\leq& \frac{1}{n}(x-y)^T(16\alpha^2I+8\alpha\beta I+\beta^2I)(x-y)\\ 
=&\frac{1}{n}\left((\sqrt{n}L-n\mu)^2+2(\sqrt{n}L-n\mu)n\mu+(n\mu)^2\right)\|x-y\|_2^2 \\
=&L^2\|x-y\|_2^2,
  \end{aligned}
\end{equation*}
which implies $\{f_i\}_{i=1}^n$ is $L$-average smooth.

\subsubsection{Proof of Lemma \ref{lemma:minimizer_case2}}
The first order optimality condition for \autoref{obj2} is that
\begin{equation*}
\begin{aligned}
\left(\frac{\sqrt{n}L-n\mu}{4}\sum_{i=1}^{n}A_i+n\mu\sum_{i=1}^{n}B_i \right)x \\
-\frac{\sqrt{n}L-n\mu}{4}\sum_{i=1}^{n}e_{(i-1)p+1}=0.
\end{aligned}
\end{equation*}
Rewriting this equation, we have
\begin{equation*}
\sum_{i=1}^{n}\left(A_i+\frac{4\sqrt{n}}{\kappa-\sqrt{n}} B_i \right)x=\sum_{i=1}^{n}e_{(i-1)p+1},
\end{equation*}
where $\kappa=\frac{L}{\mu}$.

Let $x=(x_{1,1},x_{1,2},\ldots,x_{1,p},x_{2,1}, x_{2,2},\ldots,x_{2,p},\ldots, \\ x_{n,1},x_{n,2}\ldots,x_{n,p})^T$, then for any $i\in\{1,2,\ldots,n\}$, we obtain
\begin{equation*}
  \begin{aligned}
\frac{2(\kappa+\sqrt{n})}{\kappa-\sqrt{n}}x_{i,1}-x_{i,2} &=1, \\
x_{i,k+1}-\frac{2(\kappa+\sqrt{n})}{\kappa-\sqrt{n}}x_{i,k}+x_{i,k-1} &=0, \\
\left(\xi+\frac{4\sqrt{n}}{\kappa-\sqrt{n}}\right) x_{i,p}-x_{i,p-1} &=0,
  \end{aligned}
\end{equation*}
where $k=2,3\ldots,p-1$.
Recalling that $q_i=\frac{\sqrt{\kappa}-n^{\frac{1}{4}}}{\sqrt{\kappa}+n^{\frac{1}{4}}}$, it is the smallest root of equation
\begin{equation*}
q_i^2-\frac{2(\kappa+\sqrt{n})}{\kappa-\sqrt{n}}q_i+1=0,
\end{equation*}
and the root of equation
\begin{equation*}
\left(\xi+\frac{4\sqrt{n}}{\kappa-\sqrt{n}}\right)q_i-1=0.
\end{equation*}
Then $x^*=(q_1,q_1^2, \ldots,q_1^p,q_2,\ldots, q_n^p)^T$ satisfies the above system.

\subsubsection{Proof of Lemma \ref{lemma:log_case2}}
Using the fact that $\log(1+x)\leq x$, we have
\begin{equation*}
  \begin{aligned}
\frac{1}{\log\frac{1}{q_1}} &=\frac{1}{\log\frac{\sqrt{\kappa}+n^{\frac{1}{4}}}{\sqrt{\kappa}-n^{\frac{1}{4}}}}=\frac{1}{\log(1+\frac{2n^\frac{1}{4}}{\sqrt{\kappa}-n^{\frac{1}{4}}})}    \\
&\geq\frac{\sqrt{\kappa}-n^{\frac{1}{4}}}{2n^\frac{1}{4}}  \\
&\geq \frac{n^{-\frac{1}{4}}\sqrt{\kappa}-1}{2}.
  \end{aligned}
\end{equation*}
Due to the relation that $x-1\geq\frac{1}{6}x+\frac{1}{4}$ when $x\geq\frac{3}{2}$, we obtain
\begin{equation*}
\frac{1}{\log\frac{1}{q_1}}\geq \frac{1}{12}n^{-\frac{1}{4}}\sqrt{\kappa}+\frac{1}{8}.
\end{equation*}
Noting that $n^{-\frac{1}{4}}\sqrt{\kappa}\geq\frac{3}{2}$, it indicates $\kappa\geq\frac{9}{4}\sqrt{n}$. This finishes the proof of the lemma. 

\end{document}